\documentclass[10pt,letterpaper]{article}
\usepackage[top=0.85in,left=2.75in,footskip=0.75in]{geometry}

\usepackage{amsmath,amssymb}

\usepackage{changepage}

\usepackage[utf8x]{inputenc}

\usepackage{textcomp,marvosym}

\usepackage{cite}
\usepackage{subfigure}
\usepackage{nameref,hyperref}

\usepackage[right]{lineno}

\usepackage{microtype}
\DisableLigatures[f]{encoding = *, family = * }

\usepackage[table]{xcolor}

\usepackage{array}

\newcolumntype{+}{!{\vrule width 2pt}}

\newlength\savedwidth

\newcommand\thickhline{\noalign{\global\savedwidth\arrayrulewidth\global\arrayrulewidth 2pt}%
\hline
\noalign{\global\arrayrulewidth\savedwidth}}

\raggedright
\usepackage[aboveskip=1pt,labelfont=bf,labelsep=period,justification=raggedright,singlelinecheck=off]{caption}

\makeatletter
\renewcommand{\@biblabel}[1]{\quad#1.}
\makeatother

\usepackage{lastpage,fancyhdr,graphicx}
\usepackage{epstopdf}
\fancyheadoffset[L]{2.25in}
\fancyfootoffset[L]{2.25in}

\newcommand{\lorem}{{\bf LOREM}}
\newcommand{\ipsum}{{\bf IPSUM}}

\begin{document}
\vspace*{0.2in}

\begin{flushleft}
{\Large
\textbf\newline{Assessment of Differentially Private Synthetic Data for Utility and Fairness in End-to-End Machine Learning Pipelines for Tabular Data} 
}
\newline
\\
Mayana Pereira\textsuperscript{1,2,*},
Meghana Kshirsagar\textsuperscript{1},
Sumit Mukherjee\textsuperscript{3},
Rahul Dodhia\textsuperscript{1},
Juan Lavista Ferres\textsuperscript{1},
Rafael de Sousa\textsuperscript{2}
\\
\bigskip
\textbf{1} AI for Good Research Lab, Microsoft, Redmond, Washington, U.S.A.
\\
\textbf{2} Department of Electrical Engineering, University of Brasilia, Brasilia, Brazil
\\
\textbf{3} INSITRO, San Francisco, CA, U.S.A.
\\
\bigskip



* mayana.wanderley@microsoft.com

\end{flushleft}
\section*{Abstract}

Differentially private (DP) synthetic data sets are a solution for sharing data while preserving the privacy of individual data providers. Understanding the effects of utilizing DP synthetic data in end-to-end machine learning pipelines impacts areas such as health care and humanitarian action, where data is scarce and regulated by restrictive privacy laws. In this work, we investigate the extent to which synthetic data can replace real, tabular data in machine learning pipelines and identify the most effective synthetic data generation techniques for training and evaluating machine learning models. We systematically investigate the impacts of differentially private synthetic data on downstream classification tasks from the point of view of utility as well as fairness. Our analysis is comprehensive and includes representatives of the two main types of synthetic data generation algorithms: marginal-based and GAN-based.

To the best of our knowledge, our work is the first that: (i) proposes a training and evaluation framework that does not assume that real data is available for testing the utility and fairness of machine learning models trained on synthetic data; (ii) presents the most extensive analysis of synthetic data set generation algorithms in terms of utility and fairness when used for training machine learning models; and (iii) encompasses several different definitions of fairness. 

Our findings demonstrate that marginal-based synthetic data generators surpass GAN-based ones regarding model training utility for tabular data. Indeed, we show that models trained using data generated by marginal-based algorithms can exhibit similar utility to models trained using real data. Our analysis also reveals that the marginal-based synthetic data generator MWEM PGM can train models that simultaneously achieve  utility and fairness characteristics close to those obtained by models trained with real data. 


\section*{Introduction}

Differential privacy (DP) is the standard for privacy-preserving statistical summaries \cite{dwork2006calibrating}. Companies such as Microsoft \cite{pereira2021us}, Google \cite{2020google}, Apple \cite{tang2017privacy}, and government organizations such as the US Census \cite{abowd2018us}, have successfully applied DP in machine learning and data sharing scenarios. The popularity of DP is due to its strong mathematical guarantees. Differential Privacy guarantees privacy by ensuring that the inclusion or exclusion of any particular individual does not significantly change the output distribution of an algorithm. 


In areas ranging from health care, humanitarian action, education, and socioeconomic studies, the publication and sharing of data is crucial for informing society and scientific collaboration. However, the disclosure of such data sets can often reveal private, sensitive information. Privacy-preserving data publishing aims at enabling such collaborations while preserving the privacy of individual entries in the data set. Tabular/categorical data about individuals are relevant in many applications, from health care to humanitarian action. Privacy-preserving data publishing for such data can be done in the form of a synthetic data table that has the same schema and similar distributional properties as the real data. The aim here is to release a perturbed version of the original information, so that it can still be used for statistical analysis, but the privacy of individuals in the database is preserved.

The biggest advantage of synthetic data sets is that, once released, all data analysis and machine learning tasks are performed in the same way it is done with real data. As noted by \cite{qian2023synthetic}, the switch between real and synthetic data in data analysis and machine learning pipelines is seamless - the same analysis tools, libraries and algorithms are applied in the same manner in both data sets. Other privacy-preserving technologies, such as federated learning, requires expertise and appropriate tools to perform data analysis and model training.


Due to the all the potential benefits of synthetic data, understanding the impacts of synthetic data in downstream classification tasks have become of extreme importance. A trend observed in recent studies is to evaluate performance of synthetic data generators of two types: marginal-based synthesizers \cite{movahedi2023evaluating} and generative adversarial networks (GAN) based synthesizers \cite{cheng2021can, ganev2021dp, qian2023synthetic}. Marginal-based synthetic data generators are suitable for tabular data only, and have gained increased popularity after the algorithm MST won the NIST competition in 2018 \cite{mckenna2021winning}. Marginal-based synthesizers are named as such due to the fact that they learn approximate data distributions by querying noisy marginals from the real data. Notable marginal-based algorithms are MWEM PGM \cite{mckenna2019graphical} and PrivBayes \cite{zhang2017privbayes}. GAN-based synthesizers, on the other hand, are flexible algorithms, and are suitable for tabular, image and other data formats. GANs learn patterns and relationships from the input data based on a game, in the sense of game theory, between two machine learning models, a discriminator model and the generator model. Among popular differentially private GAN architectures we list DP-GAN \cite{xie2018differentially}, DP-CTGAN \cite{QUAIL} , PATE-GAN \cite{jordon2018pate} and PATE-CTGAN \cite{QUAIL}.

One of the major applications of synthetic data is for training machine learning models. Therefore, it is paramount to understand how exchanging real data for synthetic data impacts the performance of the trained machine learning models. By performance, we mean not only the utility of the model (its accuracy, for example) but also how well the model performs for different subgroups of the data set - the fairness of the model. The impact of machine learning models on minorities subgroups is an active area of research, and several works have investigated the trade-offs among model accuracy, bias, and privacy \cite{wiens2019no,vitaly, calmon2017optimized,rajotte2021reducing}. However, only recently bias caused by the use of synthetic data in downstream classification received attention \cite{ganev2022robin, movahedi2023evaluating, giles2022faking}. 
This problem becomes particularly relevant in the context of synthetic data sets generated with differential privacy guarantees. It is known that differential privacy can affect fairness in machine learning models \cite{vitaly}. Despite recent work investigating the impact of synthetic data in downstream model fairness \cite{ganev2022robin, cheng2021can}, there are important questions that remain unanswered. 

\begin{itemize}

\item There is no published work that systematically studies the utility and fairness of machine learning models trained on several GAN based and marginal-based synthetic tabular data set generation algorithms. 

\item Previous studies have not evaluated machine learning models trained on synthetic data set generation algorithms for multiple definitions of fairness. 

\item In previous studies, it was always assumed that real data was available for evaluating the fairness of models trained on synthetic data. Here, we propose and evaluate a pipeline where no such assumption is necessary. 

\end{itemize}

\textbf{Contributions} In this work, we investigate the impacts of differentially private synthetic data on downstream classification, where we focus on understanding the impacts on model utility and fairness. Our investigation focus on two aspects of such impact: 

\begin{itemize}
    \item What is the impact in model utility when utilizing synthetic data for training machine learning models? Can synthetic data also be used to evaluate utility of machine learning models?
    \item What is the impact in model fairness when utilizing synthetic data for training machine learning models? Can synthetic data be used to evaluate fairness of machine learning models?
\end{itemize}

In our investigations we also evaluate if there are clear differences in performance between marginal-based and GAN-based synthetic data, and if there is a synthesizer algorithm that produces data that clearly outperform others.


\begin{figure*}[htp]
  \centering
 
 \includegraphics[width=13cm]{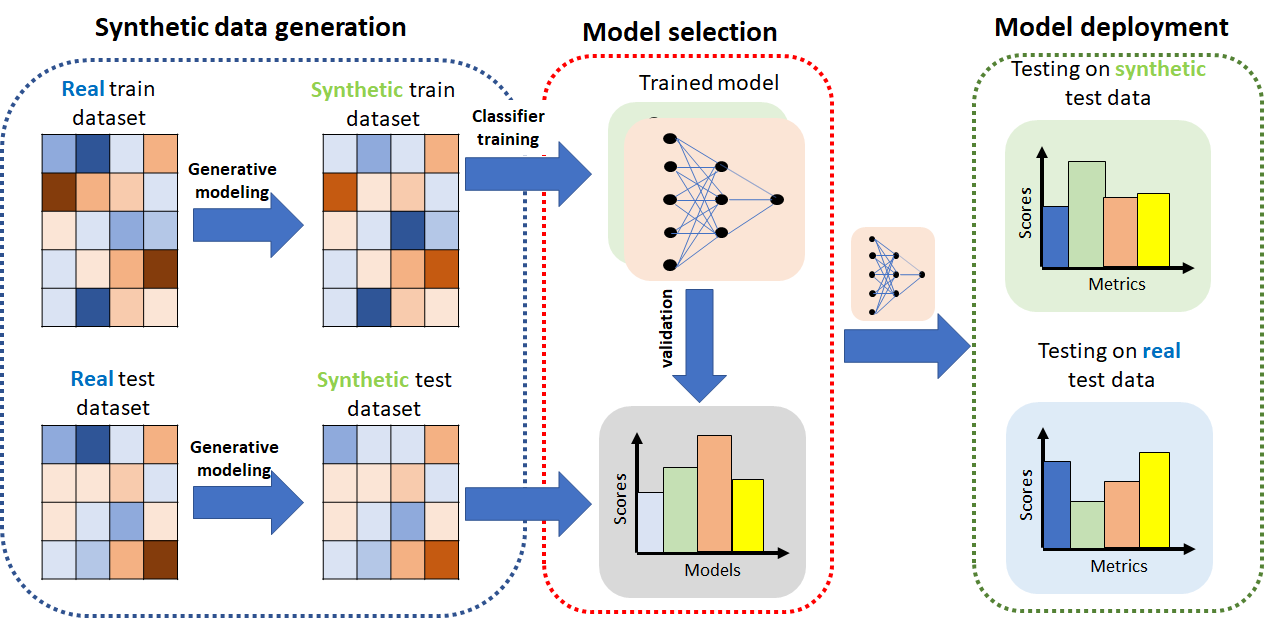}
  \caption{Pipeline for model training and evaluation using synthetic data  (1) We generate Synthetic data sets for model training and model testing utilizing differentially private synthesizers. (2) We train models utilizing synthetic data and evaluate on a synthetic test data. Model selection is made during this phase. (3) Based on the previous phase results, model is trained using synthetic data and deployed. Model is applied to real (test) data in production phase.}
   \label{img:method}

\end{figure*}

Our research work evaluates the impact of utilizing synthetic data sets for both training and testing in machine learning pipelines. We empirically compare the performance of marginal-based synthesizers and GAN-based synthesizers within the context of a machine learning pipeline. Our experiments yield a comprehensive analysis, encompassing utility and fairness metrics.
Our main contributions are: 

\begin{itemize}
    \item We propose a training and evaluation framework that does not assume that real data is available for testing the utility and fairness of machine learning models trained on synthetic data. 
    \item We present an extensive analysis of synthetic data set generation algorithms in terms of utility and fairness when used for training machine learning models. In particular, this is the first systematic comparison of several marginal-based and GAN-based algorithms for fairness and utility of the resulting machine learning models. 
    \item This is the first of such studies that includes several different definitions of fairness. 
\end{itemize}

\textbf{Main Findings:}

\begin{itemize}
    \item[1] \textbf{Marginal-based synthetic data can accurately train machine learning models for tabular data.} Marginal-based synthetic data can train models with similar utility to models trained on real data. Our experiments show that for a privacy-loss parameter $\epsilon > 5.0$, models trained with MWEM PGM (AUC = 0.684), MST (AUC = 0.662) and Privbayes (AUC = 0.668) provides utility very similar to models trained on real data (AUC = 0.684). Additionally, we evaluated models using synthetic data, and found that marginal-based synthetic provides a good evaluation, with synthetic data providing an AUC = 0.671 versus AUC = 0.684 (measured using real data).

    \item[2] \textbf{Synthetic data sets trained with MWEM PGM can be used for accurate model training and fairness evaluation in the case of tabular data.} We found that MWEM PGM synthetic data can train models that achieves very similar utility and fairness characteristics of models trained with real data. Additionally, the synthetic data generated by MWEM PGM algorithm showed very similar behavior to real data when used to evaluate utility an fairness of machine learning models. This is the first study that (first time that it is showing that synthetic data can actually present reliable behavior and a potential substitute for real data sets in end-to-end machine learning pipelines)

\end{itemize}

This work significantly extends and sub sums a previous version, presented at the \emph{Machine Learning for Data: Automated Creation, Privacy, Bias Workshop} at the \emph{International Conference on Machine Learning (ICML)} (workshop without proceedings) \cite{pereira2021analysis}.




\section{Related Works} 
As synthetic data generation becomes standard practice for data sharing and publishing, understanding the impacts of utilizing synthetic data in machine learning pipelines is of significant importance. Although previous works have advised against using synthetic data to train and evaluate any final tools deployed in the real world \cite{jordon2022synthetic}, in very sensitive scenarios, such as human trafficking data \cite{msrdata}, synthetic data might be the only available data for training and testing models. 

The promises synthetic data brings generated an interest in understanding impacts of utilizing synthetic in data analysis and machine learning. Some of these works include analysing the utility of differentially private synthetic data in different tasks \cite{tao2021benchmarking}, investigating if training models with differentially private synthetic images can increase subgroup disparities \cite{cheng2021can}, the impacts different types of synthetic data can have in model fairness \cite{ganev2022robin, bullwinkel2022evaluating}, utility of synthetic data in downstream health care classification systems \cite{movahedi2023evaluating}, and whether feature importance can be accurately analyzed using differentially private synthetic data \cite{giles2022faking}. All these works are ultimately trying to answer a same question: to which extent can we substitute real data with synthetic data, and which are the best synthetic data generation techniques for model training? 

However these works still left questions unanswered. First of all, there hasn't been a systematic study of impacts of using synthetic data sets in end-to-end machine learning pipelines, which means evaluating the use of synthetic data for model training and model evaluation. Additionally, there has been a lot of focus on image classification tasks \cite{cheng2021can, ganev2022robin} where the disparity in accuracy are largely attributable to the class imbalance in these data sets: i.e disadvantaged classes are also rare classes in the data set thereby leading to worse performance on these. In contrast, our work studies these issues in the context of tabular data sets and in settings where the data has an intrinsic bias against sub-populations that are not necessarily rare in the data set. Moreover, our work focus on comparing two types of data synthetization algorithm families: marginal-based and GAN-based data synthesizers. While, these two type of data synthetization algorithms have been previously compared for utility  \cite{tao2021benchmarking}, no such extensive comparative analysis exists for fairness. 

We are the first to extensively study the differences of applying data generated by these two families types of data synthetization algorithms in end-to-end machine learning pipelines for utility and multiple fairness metrics.

\section{Preliminaries}
In this section we introduce the concepts of differential privacy and algorithmic fairness. We refer the reader to \cite{dwork2006calibrating, heidari, barocas} for detailed explanation of these concepts. Additionally, we describe the synthetic data generation techniques and the data sets used in our experiments.
\subsection{Differential privacy}
Differential privacy is a rigorous privacy notion used to protect an individual’s data in a data set disclosure. We present in this section notation and definitions 
that we will use to describe our privatization approach.  We refer the reader to \cite{book}, \cite{mcsherry} and \cite{calibrate} for detailed explanations of these definitions and theorems. 

\textsc{Pure Differential Privacy.} A randomized mechanism $\mathcal{M}:\mathcal{D}\rightarrow \mathcal{A}$ with data base domain $\mathcal{D}$ and output set $\mathcal{A}$ is $\epsilon$-differentially private if, for any output $A \subseteq \mathcal{Y}$ and neighboring databases $D, D' \in \mathcal{D}$ (i.e., $D$ and $D'$ differ in at most one entry), we have
\[
    \textnormal{Pr}[\mathcal{M}(D) \in A] \leq e^{\epsilon}\textnormal{Pr}[\mathcal{M}(D') \in A]
\]

\textsc{Approximate Differential Privacy.} A randomized mechanism $\mathcal{M}:\mathcal{D}\rightarrow \mathcal{A}$ with data base domain $\mathcal{D}$ and output set $\mathcal{A}$ is $(\epsilon,\delta)$-differentially private if, for any output $A \subseteq \mathcal{Y}$ and neighboring databases $D, D' \in \mathcal{D}$ (i.e., $D$ and $D'$ differ in at most one entry), we have
\[
    \textnormal{Pr}[\mathcal{M}(D) \in A] \leq e^{\epsilon}\textnormal{Pr}[\mathcal{M}(D') \in A] + \delta
\]

The privacy loss of the mechanism is defined by the parameter $\epsilon \geq 0$   in the case of 'pure' differential privacy and parameters $\epsilon, \delta \geq 0$ in the case of 'approximate' differential privacy.

The definition of neighboring databases used in this paper is user-level privacy. User-level privacy defines neighboring to be the addition or deletion of a single user in the data and all possible records of that user. Informally, the definition above states that the addition or removal of a single individual in the database does not provoke significant changes in the probability of any differentially private output. Therefore, differential privacy limits the amount of information that the output reveals about any individual.

A function $f$ (also called query) from a data set $D \in \mathcal{D}$ to a result set $ A \subseteq \mathcal{A}$ can be made differentially private by injecting random noise to its output. The amount of noise depends on the sensitivity of the query. 

\subsection{Fairness Metrics}

In this section we present the definition of two different fairness metrics: Equal Opportunity \cite{heidari} and Statistical Disparity\cite{barocas}. Given a data set $W = (X, Y', C)$ with binary protected attribute $C$ (e.g. race, sex, religion, etc), remaining decision variables $X$ and predicted outcome $Y'$, we define Equal Opportunity and Statistical Disparity as follows.

 \textsc{Equal Opportunity/ Equality of Odds} requires equal True Positive Rate (TPR) across subgroups:
\[
    \textnormal{Pr}(Y'=1|Y=1,C=0)=\textnormal{Pr}(Y'=1|Y=1,C=1)
\]

where Y' is the model output.

\textsc{Statistical Parity} requires positive predictions to be unaffected by the value of the protected attribute, regardless of true label 
\[
    \textnormal{Pr} (Y' = 1|C = 0) = \textnormal{Pr} (Y' = 1|C = 1)
\]

We follow the approach of \cite{amazon, perrone2020fair} and utilize difference in Equal Oportunity (DEO) = $|\textnormal{Pr}(Y' =1|Y =1,C=0) - \textnormal{Pr}(Y' =1|Y =1,C=1)|$ and difference in Statistical Parity (DSP) = $|\textnormal{Pr} (Y' = 1|C = 0) - \textnormal{Pr} (Y' = 1|C = 1)|$ to measure model fairness.


\subsection{Differentially Private Synthetic Data Generators.}
We use several differentially private (DP) synthetic data generators that have been specifically tailored for generating tabular data with the goal of enhancing their utility for learning tasks. We consider two broad categories of approaches: i) marginal-based methods, ii) and Generative Adversarial Network (GAN) based models.

\subsubsection{Marginal-based methods}
\paragraph{MWEM PGM} Is a variation of the multiplicative weights with exponential mechanism algorithm (MWEM), which is an algorithm that generated synthetic data based on linear queries. The algorithm aims to produce a data distribution that produces query answers similar answers resulted when querying the real data set. The MWEM PGM variation combines probabilistic graphical models with the MWEM algorithm. The structure of the graphical model is determined by the measurements, such that no information is lost relative to a full contingency table representation.

\paragraph{MST} Is a synthetic data generation algorithm that acts selecting 2- and 3-way marginals for measurement. It combines one principled step, which is to find the maximum spanning tree (MST) on the graph where edge weights correspond to mutual information between two attributes, with some additional heuristics to ensure that certain important attribute pairs are selected, and a final step to select triples while keeping the graph tree-like.

\paragraph{PrivBayes}
In order to improve the utility of the generated synthetic data, \cite{zhang2017privbayes} approximates the actual distribution of the data by constructing a Bayesian network using the correlations between the data attributes. This allows them to factorize the joint distribution of the data into marginal distributions. Next, to ensure differential privacy, noise is injected into each of the marginal distributions and the simulated data is sampled from the approximate joint distribution constructed from these noisy marginals.

\subsubsection{GAN-based methods}
Generative neural networks (GANs) are a type of artificial neural network used in machine learning for generating new data samples similar to a given training data set.
Generative adversarial networks are based on a game, in the sense of game theory, between two machine learning models, a discriminator model $D$ and the generator $G$ model. The goal of the generator is to learn realistic samples that can fool the discriminator, while the goal of the discriminator is to be able to tell generator generated samples from real ones \cite{xie2018differentially}.

\textbf{Conditional Tabular GAN (CTGAN)} \cite{xu2019modeling} is an approach for generating tabular data. CTGAN adapts GANs by addressing issues that are unique to tabular data that conventional GANs cannot handle, such as the modeling of multivariate discrete  and mixed discrete and continuous distributions. It achieves these challenges by augmenting the training procedure with mode-specific normalization, and by employing a conditional generator and training-by-sampling that allows it to explore discrete values more evenly. When applying differentially private SGD (DP-SGD) \cite{abadi2016deep} in combination with CTGAN the result is a DP approach for generating tabular data. 

The \textbf{PATE (Private Aggregation of Teacher Ensembles)} framework \cite{papernot2016semi} protects the privacy of sensitive data during training, by transferring knowledge from an ensemble of teacher models trained on partitions of the data to a student model. To achieve DP guarantees, only the student model is published while keeping the teachers private. The framework adds Laplacian noise to the aggregated answers from the teachers that are used to train the student models. CTGAN can provide  differential privacy by applying the PATE framework. We call this combination PATE-CTGAN, which is similar to PATE-GAN \cite{jordon2018pate}, for images. The original data set is partitioned into $k$ subsets and a DP teacher discriminator is trained on each subset. Further, instead of using one generator to generate samples, $k$ conditional generators are used for each subset of the data.

\subsection{Data sets}
\paragraph{Adult data set} In the Adult data set (32561 instances), the features were categorized as protected variable (C): gender (male, female); and response variable (Y): income (binary); decision variables (X): the remaining variables in the data set. We map into categorical variables all continuous variables.
 
\paragraph{Prison Recidivism data set}
From the COMPAS data set (7214 instances), we select severity of charge, number of prior crimes, and age category to be the decision variables (X). The outcome variable (Y) is a binary indicator of whether the individual recidivated (re-offended), and race is set to be the protected variable (C). We utilize a reduced set of features as proposed in \cite{calmon2017optimized}.
\paragraph{Fair Prison Recidivism data set} \label{fair_data}
We construct a "fair" data set based on the COMPAS recidivism data set by employing a data preprocessing technique for learning non-discriminating classifiers from \cite{kamiran2012data}, which involves changing the class labels in order to remove discrimination from the data set. This approach selects examples close to the decision boundary to be either 'promoted', i.e label flipped to the desirable class, or `demoted', i.e label flipped to the undesirable class (ex: the 'recidivate' label in the COMPAS data set is the undesirable class). By flipping an equal number of positive and negative class examples, the class skew in the data set is maintained.

\section{Experimental Evaluation}
One potential outcome of synthetic data sharing is the utilization of synthetic data for training and evaluating an ML model. The trained model could be deployed without assessing its performance on real data, due to lack of data access. However, it is important to acknowledge that these trained models are ultimately applied to real data. This scenario is illustrated in Figure \ref{img:method}. In our experiments, we address the concern that there may be substantial disparities in performance between the evaluation phase (employing synthetic data) and the deployment phase (utilizing real data).
We compare the performance of logistic regression models trained with differentially private synthesizers, focusing on two performance dimensions: utility and fairness. The follow the approach of \cite{ganev2022robin} and use logistic regression for downstream classification evaluation to avoid another layer of stochasticity.

To assess the utility performance, we employ the AUC-ROC metric, which quantifies trade-off between the recall and false positive rate. We examine fairness performance through three different perspectives.
Previous research \cite{vitaly} has indicated that differentially private machine learning models tend to perform worse on minority groups. To this point we evaluate the decay in accuracy for the different subgroups in the protected attribute. We also measure the difference in equality of odds (DEO) and the difference in statistical parity (DSP). These metrics allow us to assess any disparities or bias in the model's predictions across different groups. Furthermore, we also investigate the extent to which one can accurately assess a model utilizing synthetic data sets. Again, we evaluate two performance dimensions: utility and fairness. 

Our experiments include two types of synthesizers: marginal-based and GAN-based synthesizers.
We generate synthetic data using three differentially private marginal-based synthesizers: MST \cite{mckenna2021winning}, MWEM-PGM \cite{mckenna2019graphical} and PrivBayes \cite{privbayes}; and four GAN-based synthesizers: DP-GAN, DP-CTGAN, PATE-GAN and PATE-CTGAN \cite{QUAIL}. For each synthetic data generation technique, we generate data sets utilizing different four privacy-loss budgets $\epsilon = \{0.5, 1.0, 5.0, 10.0\}$.


We randomly divide the real data set into an 80/20 split, separating the data into generator and test data sets. We run 10 rounds of synthetic DP data generation on the 80\% split (generator data), where we generate synthetic train and synthetic test data sets. We utilize the SmartNoise Library\footnote{https://smartnoise.org} implementation of the synthesizers, and approximate-DP approaches use the library's default value of $\delta$. For experiments using PrivBayes Synthesizers, we use the DiffPrivLib implementation \footnote{https://github.com/IBM/differential-privacy-library}. 

We train Logistic Regression models using the generated DP synthetic data sets. In experiments where we test the trained models on real data, model performance is evaluated on the real test data (the 20\% test split from the real data). In experiments where we test the trained models on synthetic data, models are evaluated using the synthetic test data sets. 

 We report, for each technique and each value of privacy loss parameter, the mean across 10 rounds. Our experiments use three data sets: the UCI Adult data set \cite{adult} and ProPublica’s COMPAS recidivism data \cite{compass}, and a fair COMPAS data set as defined in Section \ref{fair_data}. The fair COMPAS data set provides a way to evaluate synthetic data generation performance in fair and biased versions of the same data set.

\subsection{Utility analysis of synthetic data in machine learning pipelines}

We evaluate the quality of models trained with synthetic data sets by measuring AUC and accuracy of the protected class. We consider privacy-loss budgets of 0.5, 1.0, 5.0 and 10.0 .  We compare the AUC obtained in our experiments with the AUC measured by training models with the real (non-synthetic) Adult, COMPAS, and fair COMPAS data sets.

\begin{figure}
    \centering
    \subfigure[AUC variation of models trained on synthetic data and evaluated using real data.]{\includegraphics[width=1.0\textwidth]{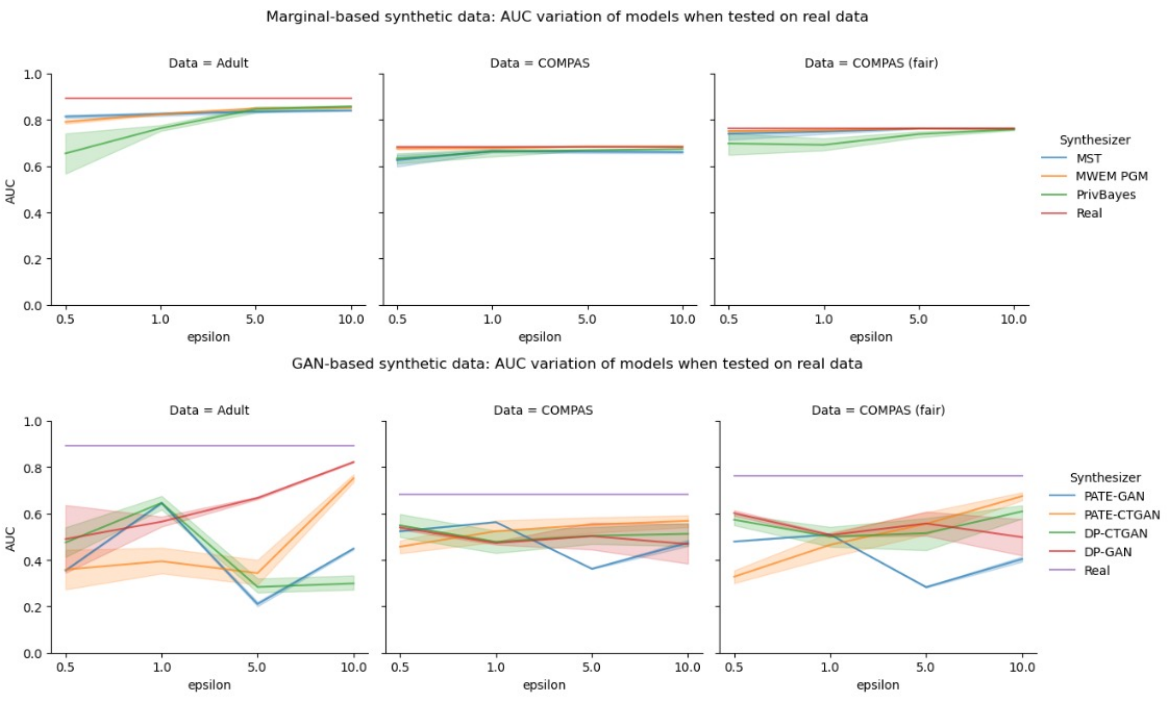}} 
    \subfigure[AUC variation of models trained on synthetic data and evaluated using synthetic data.]{\includegraphics[width=1.0\textwidth]{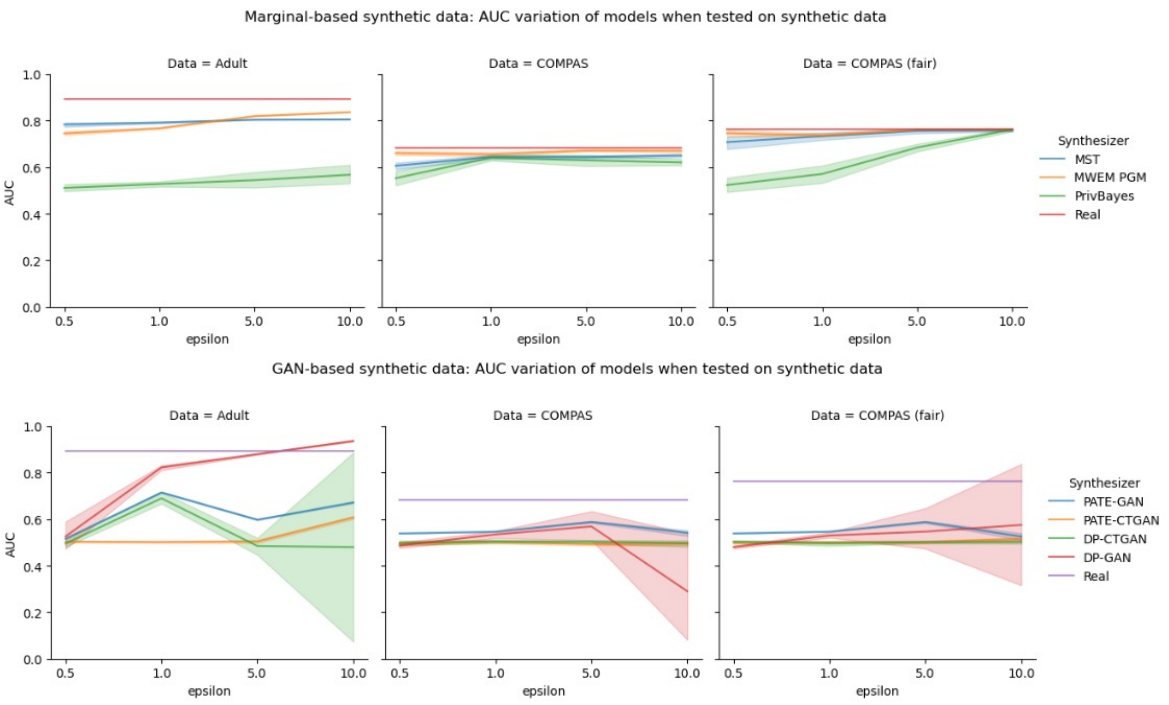}} 
    \caption{Impact in utility caused by the use of differentially private synthetic data in model training and testing. In (a) we show the decay in model utility when utilizing marginal-based and GAN-based synthetic data sets for model training. In (b) we show what is the measured model utility when the instrument for measuring model performance is a synthetic data set.}
    \label{fig:utility1}
\end{figure}

Figure \ref{fig:utility1} (a) shows AUC for different privacy losses and different synthesizers. The plots show the variation of AUC as a function of $\epsilon$ for marginal-based and GAN-based synhtesizers. The top row refers to marginal-based synthesizers. Overall, the performance of the models trained on marginal-based synthetic data is very close to the baseline model, trained on real data. For all three synthesizers, we see an increase in AUC as we increase $\epsilon$. For all data sets, Adult, COMPAS and fair COMPAS, the perfomance of MST and MWEM-PGM are similar across all values of $\epsilon$. PrivBayes has a slightly lower performance. For $\epsilon >  5.0$, all three synthesizer presented very similar performance. For COMPAS data set (which has a small dimension) the performance of synthetic data sets as training data is very close to the performance of the real data. The bottom row of figure \ref{fig:utility1} (a) presents the perfomance of GAN-based synthetic data. The overall performance of this type of synthesizer is worse and the performance of the marginal-based synthesizer. As noted by \cite{tao2021benchmarking}, models trained on GAN-based synthetic data perform worse than models trained on marginal-based synthetic data. With AUC $\approx 0.5$, we can say that they do not do much better than random guessing. Additionally, we see a much greater variance in results for a same privacy-loss budget, which is observed by the large error bars. Finally, as the privacy-loss budget increases, the utility does not necessarily increase.

Although several works have assessed the performance of machine learning models trained with synthetic data sets \cite{tao2021benchmarking, ganev2022robin, giles2022faking}, this is the first study to analyze if synthetic data sets can be used for model assessment, and how close to reality such assessment is. In Figure \ref{fig:utility1} (b) we present the plots of variation of AUC for different values of epsilon. The plots in the first line refer to performance of models trained on marginal-based synthesizers, the the plots in the second line refer to GAN-based synthesizers. By comparing the evaluation of models trained with marginal-based data in Figure \ref{fig:utility1} (a) - assessment with real data, and in Figure \ref{fig:utility1} (b) - assessment with synthetic data, we see that the assessment is very similar in both cases when the synthesizers are MST and MWEM PGM. When assessing with synthetic data, we notice that PrivBayes present a large difference in assessment results when assessing model trained on Adult and fair COMPAS synthetic data. GAN-based synthetic data present inconsistent behavior when used for model assessment. When comparing the assessments in Figure \ref{fig:utility1} (a) - assessment with real data, and (b) - assessment with synthetic data, we notice that using DP-GAN sythetic data for model assessment can over estimate model AUC. Overall, GAN-based synthetic data will make assessments that are as good as random guessing.

\paragraph{Marginal-based synthetic data does better at training and assessing utility of models.}

We ranked the utility performance of all synthesizers taking based on two criteria: ability to generate synthetic data for model training and ability to generate synthetic data for model assessment. Table \ref{AUC_comp} shows the ranking of synthesizers when generating training ans assessment data for the Adult data and the COMPAS data. The table also shows model AUC metrics when measured with real data - AUC(R), and model AUC when measured with synthetic data - AUC(S). All table results accounts for synthetic data generated with privacy-loss parameter $\epsilon = 5.0$.

\begin{table}[]  
\begin{tabular}{l cc cc cc}

 & \multicolumn{2}{c}{\textsc{Rank}}  & \multicolumn{2}{c}{ \textsc{Adult} } & \multicolumn{2}{c}{ \textsc{COMPAS} }\\ 
\textsc{Synthesizer} & \textsc{Adult} & \textsc{COMPAS} & AUC (R) & AUC (S)  & AUC (R) & AUC (S)\\ \hline \hline
MWEM PGM & 1st & 1st & 0.850 &  0.820 & 0.684 & 0.671\\
MST & 2nd & 2nd & 0.836 & 0.804 & 0.662 & 0.643\\
PrivBayes & 3rd & 3rd & 0.846 & 0.544 & 0.668 & 0.629\\
DP-GAN & 4th & 6th & 0.667 & 0.880 & 0.503 & 0.568\\
PATE-CTGAN & 5th & 4th & 0.343 & 0.504 & 0.552 & 0.492\\
DP-CTGAN & 6th & 5th & 0.284 & 0.485 & 0.504 &0.502\\
PATE-GAN & 7th & 7th & 0.210 & 0.597 & 0.362 & 0.587\\
\end{tabular}  
\caption{Synthesizer utility comparison. We compare and rank all synthesizers by their ability to generate quality training data and evaluation data for machine learning pipelines. The comparison presented accounts for synthetic data generated with privacy-loss parameter $\epsilon = 5.0$. In addition to present a performance ranking for Adult and COMPAS data, we show a comparison of model AUC measured with real data - AUC(R), and model AUC measured with synthetic data - AUC(S).}
\label{AUC_comp}
\end{table} 

MWEM PGM synthetic data outperforms all other synthetic data for both tasks: utility as training data for machine learning models and utility as evaluation data for machine learning models.  The performance of synthetic data sets generated with MWEM PGM and MST perform well and with a small performance decay when compared to real data, both when using the synthetic data for model training and model assessment. For model training, when comparing the AUC achieved by model trained with the real data set (AUC = 0.892 ) to the metrics achieved by models trained with MWEM PGM data (AUC = 0.850 ) and MST (AUC = 0.836), the decrease in performance is small. The synthetic data sets also present a good performance as assessment data. The model assessment resulted when using MST data (AUC = 0.804) and MWEM PGM data (AUC = 0.820) presents consistent results with a small decay.  Although PrivBayes data presents good performance in model training (AUC = 0.846), there is a significant discrepancy between assessment utilizing real data and assessment utilizing synthetic data. We reach similar conclusions when analysing results for COMPAS data. Using GAN-based data as training data resulted in models with utility very close to random guess, as already observed in previous analysis, with DP-GAN synthetic data performing slightly better than the rest of GAN-based data sets.


\subsection{Fairness analysis of synthetic data in machine learning pipelines}


\paragraph{Impacts on subgroup accuracy}In the previous section, we showed that adding privacy by utilizing synthetic data sets in machine learning pipelines results in a utility decrease. We now proceed to perform a fairness analysis. In this experiment, presented in Table \ref{Acc_comp} we analyze model accuracy for different groups in the protected class to understand whether the addition of privacy to the data pipeline harms model utility more for the minority class than it does for the privileged class. Results in Table \ref{Acc_comp} refer to the Adult data set.

\begin{table}[]  
\begin{tabular}{l cc cc}  
 &  \multicolumn{4}{c}{\textsc{Accuracy of different subgroups - Adult data}}\\  
\textsc{Synthesizer} & minority (R) & minority (S) & privileged (R) & privileged (S)\\  \hline \hline
 Real & 0.924 & -- & 0.804 & --\\  
MWEM PGM &  0.909 & 0.898 & 0.779 & 0.770 \\  
MST &  0.914 & 0.895 & 0.756 & 0.765 \\   
PrivBayes & 0.892 & 0.596 & 0.709  & 0.575\\  
DP-GAN &  0.733 & 0.929 & 0.585 & 0.855\\  
PATE-CTGAN  & 0.892 & 0.938 & 0.695 & 0.942 \\  
DP-CTGAN & 0.889 & 0.999 & 0.693 & 0.999\\  
PATE-GAN & 0.892 & 0.874 & 0.695 & 0.854\\  
\end{tabular}  
\caption{Accuracy comparison for different groups. The comparison presented accounts for synthetic data generated with privacy-loss parameter $\epsilon = 5.0$. We show a comparison of model accuracy for the different groups measured with real data (R), and model accuracy measured with synthetic data (S).}
\label{Acc_comp}
\end{table}  
From a fairness perspective, the overall behavior of all synthesizers is to have less accuracy decay for the protected class than it does for the privileged class. As observed on the utility experiments, MWEM PGM and MST are the best performing synthetic data sets for both pipeline tasks: training and evaluation. Although MWEM PGM presents good results for minority and privileged classes, where the model accuracy is very close to the baseline model - captured by accuracy minority(R) and accuracy privileged(R) in Table \ref{Acc_comp}. Additionally, evaluation with MWEM PGM synthetic data sets captured accuracy metric for both classes - captured by accuracy minority(S) and accuracy privileged(S) - that are very close to model evaluation done with real data.

\paragraph{Impacts on statistical parity}

A model presents statistical parity if the percentage of positive predictions are the same for all subgroups. The goal of the experiments in this section is to measure whether models trained with synthetic data preserve the  characteristics of models trained on real data. 

Our experiments measure the difference in statistical parity (DSP) of models. We measure DSP of models using real data - DSP(R), and using synthetic data - DSP(S). We present a detailed comparison of DSP for all three data sets and all synthesizers on Table \ref{dsp_comp}. We notice from our experiments that several models trained on synthetic data seem to be less biased than the model trained on real data. MWEM PGM synthesizer presented the best utility overall, based on the results present in the previous experiments. PATE-CTGAN, however, was ranked in 5th place in utility. 

\begin{table}[h]  
\centering  
\begin{tabular}{llccc} 
\textsc{Data} & \textsc{Synthesizer} & DSP(R) & DSP(S) & DSP delta \\  \hline \hline

 Adult & MST & 0.083 & 0.072 & 0.011 \\  
 & MWEM PGM & 0.168 & 0.159 & 0.009 \\  
  & PrivBayes & 0.051 & 0.035 & 0.016 \\

 & DP-CTGAN & -0.001 & 0.000 & -0.001 \\  
  & DP-GAN & 0.346 & 0.253 & -0.093 \\  
  & PATE-CTGAN & 0.000 & 0.000 & 0.000 \\  
  & PATE-GAN & 0.000 & 0.000 & 0.000 \\  
  & Real & \textbf{0.189} \\ 
  
 COMPAS & MST & -0.182 & -0.101 & -0.082 \\  
 & MWEM PGM & -0.218 & -0.190 & -0.028 \\  
  & PrivBaeys & -0.211 & -0.166 & -0.046 \\  
  
  & DP-CTGAN & -0.034 & 0.001 & -0.034 \\  
 & DP-GAN & 0.072 & -0.089 & 0.161 \\  
  & PATE-CTGAN & -0.008 & -0.009 & 0.001 \\  
 & PATE-GAN & 0.000 & -0.001 & 0.001 \\  
  & Real & \textbf{-0.205}  \\  
  
COMPAS & MST & -0.185 & -0.090 & -0.095 \\  
 (fair)& MWEM PGM & -0.018 & 0.015 & -0.032 \\  
 & PrivBayes & -0.065 & 0.037 & -0.027 \\

 & DP-CTGAN & -0.034 & -0.004 & -0.030 \\  
 & DP-GAN & 0.066 & 0.096 & -0.030 \\  
 & PATE-CTGAN & 0.000 & 0.000 & 0.000 \\  
 & PATE-GAN & 0.000 & 0.000 & 0.000 \\  
 & Real & \textbf{-0.025}  \\  
 
\end{tabular} \caption{Difference in statistical parity (DSP) of models trained with synthetic data. We measure the DSP of models using real test data - DSP(R) and synthetic test data DSP(S). DEO delta quantifies the difference between DSP(R) and DSP(S). All synthetic data where generated using privacy-loss parameter $\epsilon = 5.0$.} 
\label{dsp_comp}
\end{table}  

To understand better what is behind this apparent fairness provided by PAET-CTGAN, we investigate the percentage of positive labelled samples in the training data, evaluation data and predictions. We present percentages for minority and privileged classes for adult data in Table \ref{pos_labels}.
\begin{table}[]  
\begin{tabular}{l cc cc cc } 
& \multicolumn{6}{c}{\textsc{Ratio of positive labels - Adult data}}
\\
\textsc{Generation} & \multicolumn{2}{c}{\textsc{Generated data}}& \multicolumn{2}{c}{\textsc{Predictions(R)}} & \multicolumn{2}{c}{\textsc{Predictions(S)}}\\ 

\textsc{algorithm} & \textsc{Female} & \textsc{Male} & \textsc{Female} & \textsc{Male} & \textsc{Female} & \textsc{Male}\\

\hline \hline
Real & 0.109 & 0.303  & 0.055 & 0.244 &\\

MWEM PGM & 0.120 & 0.307 & 0.042 & 0.209 & 0.043 & 0.202\\
MST & 0.123 & 0.297 & 0.032 & 0.115 & 0.031 & 0.102\\
PrivBayes & 0.265 & 0.342 & 0.004 & 0.055 & 0.056 & 0.091\\
PATE-GAN & 0.125 & 0.144 & $\approx$ 0 & $\approx$ 0 & $\approx$ 0 & $\approx$ 0 \\ 
PATE-CTGAN & 0.056 & 0.058 & $\approx$ 0 & $\approx$ 0 & $\approx$ 0 & $\approx$ 0 \\ 
DP-GAN & 0.061 & 0.307 & 0.199 & 0.545  & 0.016 & 0.269 \\ 
DP-CTGAN & $\approx$ 0 & 0.002 & 0.227 & 0.130  & $\approx$ 0 & $\approx$ 0 \\ 

\end{tabular}  
\caption{Ratio of samples with positive labels for each subgroup in the protect class in the Adult data. We compare percentages present in the true labels of the real data and the predicted labels. Analogously, we measure the percentage of samples with positive present in the training, testing and predicted labels for data sets generated from three distinct synthesizer techniques: MWEM PGM,  PATE-CTGAN and DP-GAN. Predictions(data1/data2) represents prediction labels of an experiment where model was trained with data1, and predictions were performed on data2.}
\label{pos_labels}
\end{table}  

We observe in Table \ref{pos_labels} that synthetic data generated with PATE-CTGAN presents a very similar percentages of samples with positive labels, of $\approx 5\%$ for each group that belongs to the protected attribute. At a first sight, this seems like a data set with promising fairness capabilities. However, when training models with such data, there are no positive predictions resulting from the model scoring. The model trained with PATE-CTGAN data acts like a majority baseline classifier for all groups. The data sets generated with DP-CTGAN presented an accentuated disparity in positive labels percentages between minority and privileged classes. In the real data 30\% of privileged class contains positive labels, while only 10\% of minority class contains positive labels. Although DP-GAN synthesizer generates data where 31\% of privileged class with positive labels (a value similar to the one presented in the real data - 30\%), there is a significant decrease in the percentage of positive class in the minority class, which is $\approx 6\%$. This imbalance is even further accentuated by the models trained with DP-GAN synthetic data. Model predictions resulted in over half of samples from the privileged class being classified with positive labels (versus 20\% of minority class).

MWEM PGM once again was the best overall performing model, as it preserves similar percentages of positive labels for all groups, 11\% and 30\% (compared to 11\% and 30\% in real data). Models trained with MWEM PGM also presented similar metric to models trained with real data, and even presenting slightly improvement in fairness.

The DSP delta presented in Table \ref{dsp_comp} quantifies the difference in DSP observed during model evalution with real data and model evaluation with synthetic data. For Adult data set, a positive DSP delta means that evaluation with synthetic data observed fairer 
results than evaluation with real data. For COMPAS and fair COMPAS data, a negative DSP delta means that evaluation with synthetic data observed fairer 
results than evaluation with real data. 

Across all data sets, models trained with MWEM PGM presented DSP metrics very similar to models trained with real data, this is captured by the DSP(R) metric.

\paragraph{Impacts on equal opportunity}

Equal Opportunity requires equal True Positive Rate
(TPR) across subgroups. Difference in equal opportunity (DEO) measures the difference of privileged group TPR and minority group TPR. 

We perform a thorough analysis to understand two points.First, what is the DEO of models trained with synthetic data sets, and how does it compare with models trained with real data? Second, we investigate whether synthetic data preserves similar true positive rates across all subgroups.

We present in Table \ref{deo_comp} experiment results comparing DEO of models trained with differentially private synthetic data sets ($\epsilon = 5.0$). These experiment are similar to the statistical parity experiments, we use real data - DEO(R) - to measure DEO of models trained on synthetic data, as well as synthetic data - DEO(S).
\begin{table}[]  
\begin{tabular}{ll ccc}  
\textsc{Data} & \textsc{Synthesizer} & DEO (R) & DEO (S) & DEO Delta \\  \hline \hline
Adult & MST & 0.038 & 0.076 & -0.037 \\  
 & MWEM PGM & 0.206 & 0.200 & 0.006 \\  
 & PrivBayes & 0.094 & 0.030 & 0.063 \\  

& DP-CTGAN & -0.002 & $\approx$0.00 & -0.002 \\  
 & DP-GAN & 0.527 & 0.641 & -0.116 \\  
 & PATE-CTGAN & 0.000 & 0.000 & 0.000 \\  
 & PATE-GAN & 0.000 & 0.000 & 0.000 \\ 
&	Real	& \textbf{0.173}	\\

COMPAS & MST & -0.150 & -0.089 & -0.061 \\  
& MWEM PGM & -0.215 & -0.224 & 0.009 \\  
 & PrivBayes & -0.177 & -0.158 & -0.020 \\ 
& DP-CTGAN & -0.031 & -0.000 & -0.031 \\  
 & DP-GAN & -0.075 & 0.020 & 0.055 \\  
 & PATE-CTGAN & -0.011 & -0.009 & -0.002 \\  
 & PATE-GAN & 0.000 & -0.001 & 0.001 \\  
 & Real & \textbf{-0.204} &  &  \\

COMPAS  & MST & -0.181 & -0.073 & -0.107 \\  
 (fair) & MWEM PGM & -0.019 & 0.037 & -0.056 \\  
& PrivBayes & -0.057 & 0.003 & -0.054 \\  

& DP-CTGAN & -0.030 & -0.005 & -0.026 \\  
& DP-GAN & 0.097 & 0.087 & 0.010 \\  
 & PATE-CTGAN & 0.000 & 0.000 & 0.000 \\  
 & PATE-GAN & 0.000 & -0.001 & -0.000
 \\  
 & Real & \textbf{-0.027} &  &  \\  
\end{tabular}  
\caption{Difference in equal opportunity (DEO) of models trained with synthetic data. We measure the DEO of models using real test data - DEO(R) and synthetic test data DEO(S). DEO delta quantifies the difference between DEO(R) and DEO(S). All synthetic data where generated using privacy-loss parameter $\epsilon = 5.0$.}
\label{deo_comp}
\end{table}  
The model trained with MWEM PGM synthetic data was the only one that presented a similar DEO to the baseline model, outperforming all other models trained with synthetic data. Note that our comparison, as in the DSP case, focus on understanding which synthetic data sets can train model that behave as close as possible to models trained with real data. Models trained with MST, which presented promising utility metrics and subgroup accuracy, did not capture as well the difference in equality on odds in experiments with the Adult data. For experiments with COMPAS and fair COMPAS data, MST performs better, but still worse than MWEM PGM, as we can see on Table \ref{deo_comp}.

\begin{figure}
    \centering
   \includegraphics[width=1.0\textwidth]{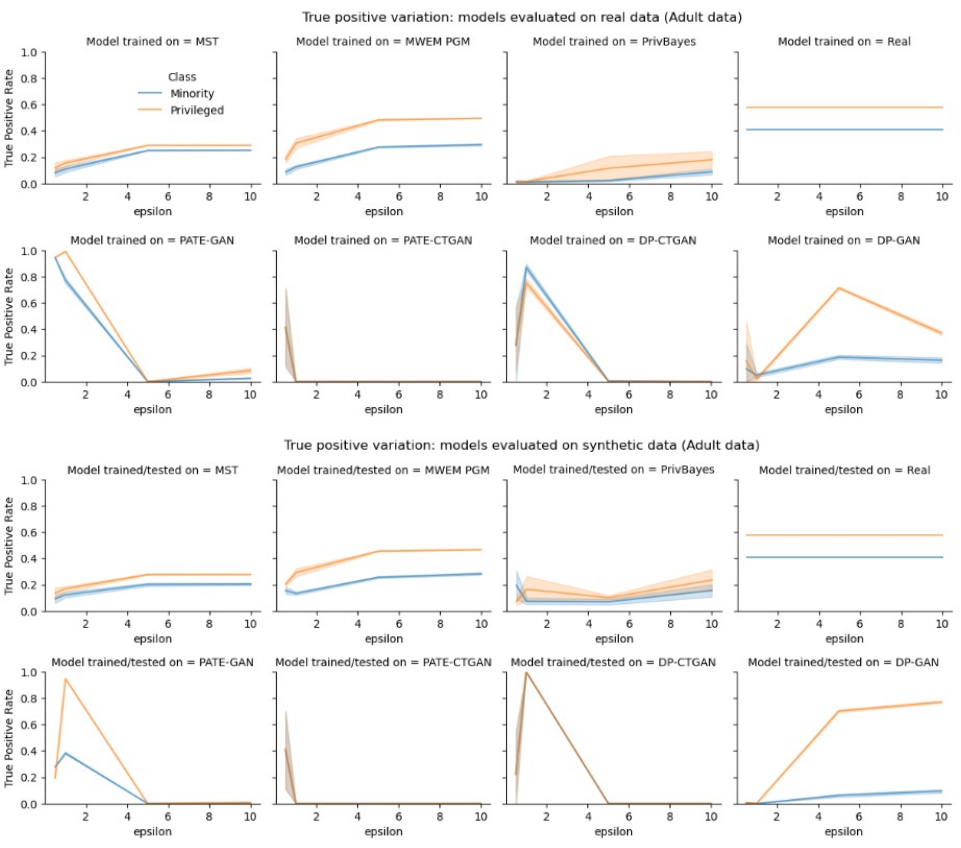}
    \caption{True positive rate (TPR) variation of different subgroups of the protected attribute of the Adult data. The top two rows shows TPR variation for different values of privacy-loss parameter $\epsilon$, of models trained with synthetic data and evaluated with real data. The bottom two rows shows TPR variation for different values of privacy-loss parameter $\epsilon$, of models trained with synthetic data and evaluated with synthetic data.}
    \label{fig:tpr1}
\end{figure}

\begin{figure}
    \centering
   \subfigure{\includegraphics[width=0.92\textwidth]{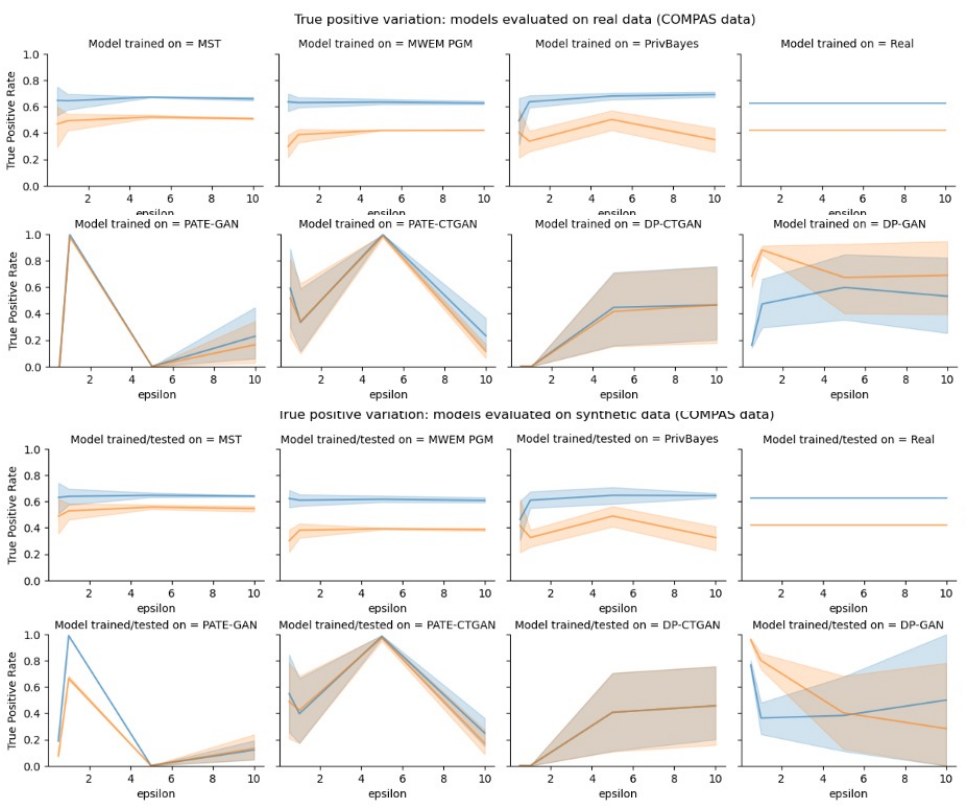}} \subfigure{\includegraphics[width=0.92\textwidth]{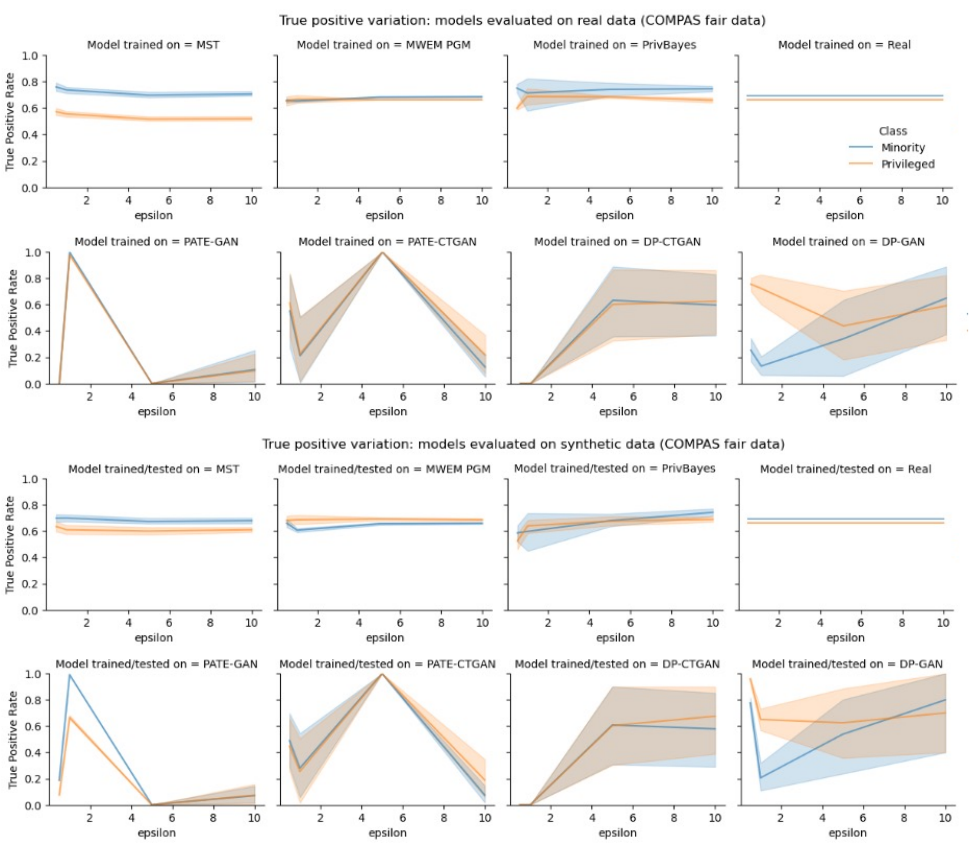}}
        \caption{TPR variation: COMPAS and $\epsilon$, of models trained with synthetic data and evaluated with real data.We also present TPR variation for different values of $\epsilon$, of models trained with synthetic data and evaluated with synthetic data.}
    \label{fig:tpr2}
\end{figure}

As we investigate the details of variation in TPR it becomes clear MWEM PGM is the the best technique for training models that preserve fairness characteristics of models trained with real data. Experiments with Adult data (Figure \ref{fig:tpr1}) show that the difference between the privileged group TPR and the minority group TPR of models trained with MWEM PGM data is very similar to the difference between subgroups TPR of models trained with real data. Experiments with COMPAS data (Figure \ref{fig:tpr2}) are even more compelling. Not only the difference between the subgroup TPR of the model trained with MWEM PGM data is close to that of the model trained with real data, but the true positive rates of the subgroups are also very similar to the TPR of the model trained with real data. Figures \ref{fig:tpr1} and \ref{fig:tpr2} show that models trained with marginal-based synthetic data outperforms models trained with GAN-based synthetic data for our tested data sets.

We make a similar analysis when evaluating how good synthetic data sets are for assessing TPRs. Figures \ref{fig:tpr1} and \ref{fig:tpr2} also present plots of TPR when synthetic data is used during model assessment. Models trained with MWEM PGM data present very similar assessment when using both real and synthetic data as test data. Models trained on MST and PrivBayes present greater discrepancies. Models trained on GAN-based data present even greater discrepancies between assessments made with real and synthetic data as test data.

\paragraph{Marginal-based synthetic data preserves and better assess model fairness}

We evaluated the performance of the synthetic data sets based on  two key model fairness tasks: the ability to mirror the behavior of actual data in downstream model fairness, and the ability to produce synthetic data for assessing model fairness. Our analysis includes a rigorous assessment of model fairness, which includes measuring subgroup accuracy, the difference in statistical parity(DSP) and the difference in equal opportunity (DEO). Beyond measuring the classical fairness metrics, we also assess the Positive Predictive Value (PPV) and True Positive Rate (TPR) for each subgroup within the protected class. The significance of evaluating PPV and TPR lies in understanding if the model upholds fairness because it accurately represents PPV and TPR for all subgroups, or if it does so merely by acting as a random classifier.

Table \ref{deo_comp2} shows the best synthesizers in end-to-end machine learning pipelines when evaluating for fairness metrics. All table results accounts for synthetic data generated with privacy-loss parameter $\epsilon = 5.0$.

MWEM PGM synthetic data, once more, outperforms all other synthetic data in the three fairness metrics. This advanatge is observed when MWEM PGM synthetic data is used as a training data set as well as when used as a testing data set.


As we investigate subgroup PPV and TPR metrics to get insights into model fairness performances. We note that MWEM PGM synthetic data presents a ratio of positive labels comparable to that obtained with real data (Table \ref{pos_labels}), for all subgroups. When evaluating the ratio of positive labels in prediction for all subgroups (female and male) in Table \ref{pos_labels}, we see that MWEM PGM also results is metrics that are the closest to real data.

The evaluation of true positive rate provides more insights into the bias introduced by synthetic data set in end-to-end machine learning pipelines. Figures \ref{fig:tpr1} and \ref{fig:tpr2} shows the variation of TPR for different values of $\epsilon$, in experiments with Adult, COMPAS ans fair COMPAS, respectively. For COMPAS data set, MWEM PGM provides performance comparable the real data set in an end-to-end analysis. For Adult data, $\epsilon > 5$ provides comparable metrics. Other algorithms, such as PrivBayes, that presented utility results (AUC metric) comparable to real data, showed low performance in terms of TPR. Finally, marginal-based synthesizers presented similar performance from the point of view of utility and fairness for both biased and fair versions of the COMPAS data set.

\begin{table}[t]  
\centering  
\begin{tabular}{l cc} 
\textsc{Metric} & \textsc{Best Synthesizer} & \textsc{Runner up} \\  \hline \hline
Subgroup accuracy & MWEM PGM & MST \\
Difference in statistical parity & MWEM PGM & MST \\
Difference in equality of odds & MWEM PGM & MST
\end{tabular}  
\caption{Difference in equal opportunity (DEO) of models trained with synthetic data. We measure the DEO of models using real test data - DEO(R) and synthetic test data DEO(S). DEO delta quantifies the difference between DEO(R) and DEO(S). All synthetic data where generated using privacy-loss parameter $\epsilon = 5.0$.}
\label{deo_comp2}
\end{table}

\section{Limitations and Future Works}
Although the data sets utilized in our analysis are commonly employed in fairness literature, extending the validity of our findings to larger-scale data sets would provide a more comprehensive understanding of the generalizability and robustness of marginal-based synthetic data approaches. Future research should focus on exploring the performance of these frameworks in real-world scenarios with diverse and extensive data sets. This would contribute to the broader applicability and reliability of synthetic data methods in various domains and facilitate a more nuanced understanding of their limitations and capabilities. Finally, extending our analysis to non-tabular data would be an interestign sequel to this work. 


\section{Conclusion}
Our research comprehensively evaluates the impact of synthetic data sets for training and testing in machine learning pipelines in the case of tabular data sets. Specifically, we compare the performance of marginal-based and GAN-based synthesizers within a machine-learning pipeline and analyze various utility and fairness metrics for tabular data sets.

Our main findings are as follows:
Marginal-based synthetic data demonstrated comparable utility to real data in end-to-end machine-learning pipelines. MWEM PGM (AUC = 0.684) provides utility very close to models trained on real data (AUC = 0.684). Furthermore, we show that model evaluation using synthetic data also provides similar results to evaluation using real data, for tabular data. The metrics obtained when utilizing marginal-based synthetic data (AUC=0.671) are comparable to real data (AUC = 0.684).
Synthetic data sets trained with MWEM PGM do not increase model bias and can provide a realistic fairness evaluation. Our study reveals that MWEM PGM synthetic data can train models that achieve similar utility and fairness characteristics as models trained with real data. Additionally, when used to evaluate the utility and fairness of machine learning models, the synthetic data generated by the MWEM PGM algorithm exhibits behavior very similar to real data. 

These findings highlight synthetic data's potential reliability and viability as a substitute for real data sets in end-to-end machine learning pipelines for tabular data. Furthermore, our research sheds light on the implications of model fairness when utilizing differentially private synthetic data for model training. 

One crucial observation is that synthetic data that does well in model training might perform differently when used as evaluation data. This was the case with Privbayes and some of the GAN-based synthetic data. This observation is important as synthetic data techniques gain acceptance as the standard data publishing approach in domains such as healthcare, humanitarian action, education, and population studies.
\bibliography{output}

\end{document}